\documentclass[journal,twoside,web]{ieeecolor}
\usepackage{jsen}
\usepackage{cite}
\usepackage{amsmath,amssymb,amsfonts}
\usepackage{algorithmic}
\usepackage{graphicx}
\usepackage{textcomp}
\usepackage{wrapfig}
\usepackage{float}  
\usepackage{subcaption} 
\usepackage{hyperref}

\usepackage{multirow}
\usepackage{picinpar}
\usepackage{multicol}
\usepackage{makecell}
\usepackage{booktabs}
\usepackage{gensymb}
\usepackage{soul}
\soulregister\cite7

\newcommand{\PAR}[1]{\vskip4pt \noindent{\bf #1~}}

\def\BibTeX{{\rm B\kern-.05em{\sc i\kern-.025em b}\kern-.08em
    T\kern-.1667em\lower.7ex\hbox{E}\kern-.125emX}}

\definecolor{abstractbg}{rgb}{0.89804,0.94510,0.83137}
\begin{document}
\title{PointLoc: Deep Pose Regressor for LiDAR Point Cloud Localization}
\author{Wei Wang, Bing Wang, Peijun Zhao, Changhao Chen, Ronald Clark, Bo Yang, \\ 
Andrew Markham, and Niki Trigoni
\thanks{This work is funded by the NIST under Grant 70NANB17H185.}
\thanks{Wei Wang, Bing Wang, Peijun Zhao, Changhao Chen, Andrew Markham, and Niki Trigoni are with the Department of Computer Science, University of Oxford, UK (e-mail: \{firstname.lastname\}@cs.ox.ac.uk). }
\thanks{Bo Yang is with the Department of Computing, The Hong Kong Polytechnic University (e-mail: bo.yang@polyu.edu.hk).}
\thanks{Ronald Clark is with the Department of Computing, Imperial College London, UK (e-mail: ronald.clark@imperial.ac.uk).}}

\IEEEtitleabstractindextext{%
\fcolorbox{abstractbg}{abstractbg}{%
\begin{minipage}{\textwidth}%
\begin{wrapfigure}[12]{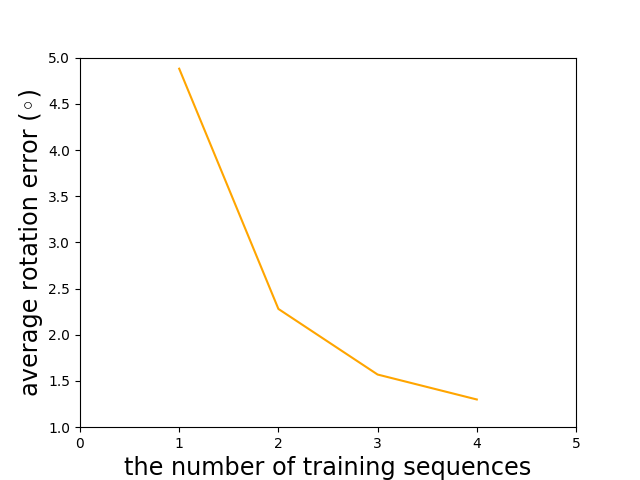}{7cm}%
\includegraphics[width=6.5cm, height=3cm]{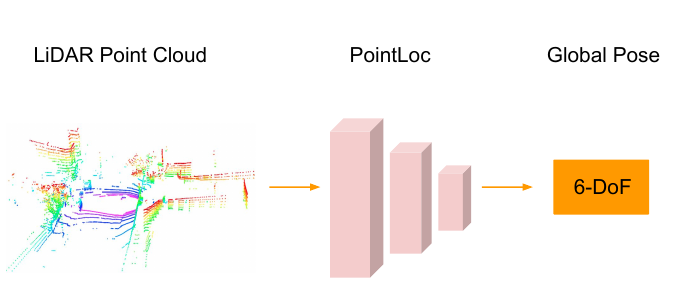}%
\end{wrapfigure}%
\begin{abstract}
In this paper, we present a novel end-to-end learning-based LiDAR sensor relocalization framework, termed PointLoc, which infers 6-DoF poses directly using only a single point cloud as input. Compared to visual sensor-based relocalization, LiDAR sensors can provide rich and robust geometric information about a scene. However, point clouds of LiDAR sensors are unordered and unstructured making it difficult to apply traditional deep learning regression models for this task. We address this issue by proposing a novel PointNet-style architecture with self-attention to efficiently estimate 6-DoF poses from $360^\degree$ LiDAR sensor frames. Extensive experiments on recently released challenging Oxford Radar RobotCar dataset and real-world robot experiments demonstrate that the proposed method can achieve accurate relocalization performance.
\end{abstract}

\begin{IEEEkeywords}
LiDAR Sensor Relocalization, LiDAR Point Cloud, Sensor Applications
\end{IEEEkeywords}
\end{minipage}}}

\maketitle

\section{Introduction}
\label{sec:introduction}
\IEEEPARstart{L}{iDAR} sensor-based localization methods have achieved impressive accuracy \cite{Kim2017RobustArea,Levinson2010}. For our motivating scenarios like indoor service robots, emergency rescue robots, and autonomous delivery service robots, several solutions have already mounted LiDARs on these robots for localization and obstacle avoidance. A typical LiDAR sensor localization system usually includes a feature extraction module, a feature matching algorithm, an outlier rejection step, a matching cost function, a spatial searching or optimization method and a temporal optimization or filtering mechanism \cite{Lu2019L3-NetDriving}. Although these geometric localization methods achieve high accuracy in some scenarios, they require significant hand-engineering efforts to tune the huge amount of hyper-parameters, and depend heavily on the running environments.

A number of novel map-based approaches have been proposed to estimate global poses \cite{Barsan2018LearningMap,Wei2019LearningMaps,Lu2019L3-NetDriving}. For example, Barsan and Wang \textit{et al.} \cite{Barsan2018LearningMap} proposed to learn descriptors from LiDAR intensity, and relocalization was performed by matching descriptors against pre-built intensity maps. However, although these methods achieve considerable accuracy, it is hard to build the map and the computing complexity increases greatly when the map gets larger. Additionally, they require other systems to provide an accurate initial pose as coarse localization first.

\begin{figure}
  \centering
  \includegraphics[width=0.8\columnwidth]{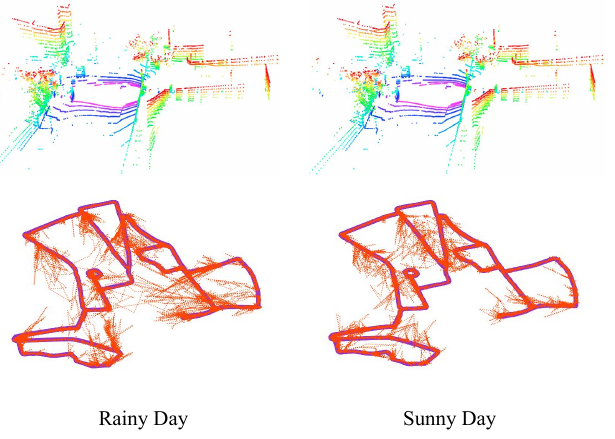}
  \caption{PointLoc results for the challenging Oxford Radar RobotCar dataset \cite{Barnes2020TheDataset,Maddern2016}. We directly feed a point cloud of the LiDAR sensor from a single timestamp to the neural network for predicting the 6-DoF pose without the requirement of pre-built maps. The  estimations of PointLoc are robust regardless of weather, and outperform the state-of-the-art DNN-based LiDAR and visual sensor relocalization methods significantly.}
  \label{fig_overview}
\end{figure}

Thus, most localization solutions employ Global Navigation Satellite System (GNSS) to provide pose estimations. Unfortunately, GNSS is not always available such as in indoor environments and the accuracy of GNSS cannot be guaranteed in areas like large cities where high-rising buildings can block the GNSS signals. To this end, Uy~\textit{et al.} \cite{Uy2018PointNetVLAD:Recognition} proposed a point cloud retrieval-based localization method to deal with the situation when the GNSS is absent. It obtains a 6-DoF pose with respect to the pre-built map in the form of reference database. Dube~\textit{et al.}~\cite{Dube2018SegMap:Descriptors} further proposed SegMap to improve the storage efficiency of the reference database by storing data-driven descriptors of individual objects in point clouds of LiDAR sensors. However, retrieval-based approaches inherently suffer from several issues. First, the time complexity of finding the closest match between the query point cloud and the reference point cloud is \textit{O}(n) where n is the number of point clouds, which is not suitable for common real-time application scenarios. Second, the point cloud-based method requires a reference database, which occupies \textit{O}(n) storage space and cannot be deployed on many mobile robots. Third, the recall rate of it is often not good enough \cite{Uy2018PointNetVLAD:Recognition}. 

Recently, learning-based approaches have emerged as a promising tool to build up a completely end-to-end localization system. These methods do not require any reference databases during runtime, and the learned features tend to be general and robust. These kind of localization approaches train a neural network to directly predict the pose. Their time complexity during inference time is \textit{O}(1) and the space it occupies is only the model size, which addresses the drawbacks of point cloud retrieval-based methods. Early attempts in this direction include PoseNet and its variations \cite{Kendall2015,Kendall2017,Clark2017a}. However, all the current pose regression approaches utilize RGB images of visual sensors as inputs, which have several problems. Visual sensors are sensitive to the change of environments, resulting in suboptimal localization performance. In addition, the input images are restricted to a narrow Field-of-View (FoV). These aspects restrict the application of these approaches to the real world. Compared with RGB images of visual sensors, point clouds, acquired by LiDAR sensors, capture $360\degree$ 3-D space, and provide much richer geometric information of a specific location. In addition, the features extracted from point clouds tend to be more robust compared to those extracted from images. However, point clouds of LiDAR sensors are unordered and unstructured, making it difficult to learn features for localization. Motivated by this, we design a neural network to use LiDAR point clouds as input for robust and accurate localization.

In this paper, we propose a novel neural network-based 3-D pose regressor, named PointLoc, to accurately estimate the 6-DoF pose using point clouds of LiDAR sensors. The neural network directly takes a primitive point cloud as input and estimates the 6-DoF pose in an end-to-end fashion. The performance shows significant improvement over the learning-based LiDAR and visual sensor relocalization methods. Fig.\ref{fig_overview} illustrates the superior performance of our PointLoc approach in different environments found in the Oxford Radar RobotCar dataset. 

In summary, our contributions are as follows:
\begin{itemize}
    \item To the best of our knowledge, this is the first LiDAR sensor-based approach for deep global pose regression in an end-to-end fashion. Our proposed architecture with a self-attention module can further improve the accuracy of the predicted 6-DoF absolute poses.
    \item We conduct real-world robot experiments in an indoor environment. We collect and create a new indoor LiDAR-visual sensors dataset, dubbed vReLoc, and release it for studying the indoor relocalization task. 
    \item Comprehensive experiments and an ablation study on these two new datasets have been done to evaluate our proposed method. Results demonstrate that the PointLoc model outperforms the state-of-the-art DNN-based LiDAR and visual sensor relocalization methods by a large margin.
\end{itemize}

The rest of the paper is organized as follows. Section~\ref{sec:related_work} introduces visual sensor relocalization, learning-based localization systems, DNN-based LiDAR odometry and deep learning on LiDAR point clouds. Problem formulation is given in Section~\ref{sec:problem_statement}. The detailed LiDAR sensor relocalization method is illustrated in Section~\ref{sec:method}. Section~\ref{sec:vReLoc} explains the collected indoor dataset vReLoc. Experiments and ablation studies are presented in Section~\ref{sec:experiments}. 
Section~\ref{sec:conclusion} concludes the work and give the future research directions.

\section{RELATED WORK}
\label{sec:related_work}
In this section, we review different learning-based approaches for localization, LiDAR odometry which estimates ego-motions between consecutive point clouds, and DNN architectures on point clouds.

\subsection{Visual Sensor Relocalization} 
For dealing with the drawbacks of map registration methods, recent works propose learning-based approaches to estimate the global pose directly \cite{Brahmbhatt2018,Kendall2015,Kendall2017,Melekhov2017Image-BasedNetworks,Clark2017a,Walch2017,Huang2019PriorEnvironments}. They take images, either single or sequential,  as inputs to train a neural network model for predicting absolute poses. The key to these methods is to learn a deep pose regressor, which usually comprises a feature extractor and a regressor \cite{Kendall2015,Huang2019PriorEnvironments,Wang2020AtLoc:Localization}. For example, PoseNet related works \cite{Kendall2015,Kendall2016Modellingrelocalization,Kendall2017} proved the feasibility of predicting the global pose using a single RGB image by regressing the pose directly. Brahmbhatt \textit{et al.} \cite{Brahmbhatt2018} utilized the relative pose between two images as a geometric constraint to estimate the pose. 
Although DNN-based relocalization methods can solve the downsides of retrieval-based approaches, the performance of translation and rotation estimation is still not satisfactory enough to be applied to real-world scenarios \cite{Sattler2019UnderstandingRegression}, which calls for further work on learning algorithms. Our work follows this line of study, aiming to improve the accuracy of deep global pose regression with LiDAR sensors.

\subsection{Learning-based Localization Systems}
Learning-based localization systems have gained significant interests recently. Almalioglu~\textit{et al.}~\cite{Almalioglu2020Milli-RIO:Radar} employed recurrent neural network for robust MMWave radar-based ego-motion estimation. CellinDeep~\cite{Rizk2019CellinDeep:Learning} adopted DNN to capture the non-linear relationship between the cellular signal and its location for robust and accurate indoor localization. Alshamaa~\textit{et al.}~\cite{Alshamaa2019DecentralizedFunctions} proposed a decentralized kernel algorithm for sensor localization in indoor wireless environments. Silva~\textit{et al.}~\cite{DaSilva2020MonocularRobots} applied transfer learning and machine learning to images of a Kinect sensor for the localization of mobile robots. Hoang~\textit{et al.}~\cite{Hoang2020Semi-sequentialEnhancement} proposed a semi-sequential probabilistic method to improve the performance of the indoor localization with extensive on-site experiments. Li~\textit{et al.}~\cite{Li2020PseudoCrowdsourcing} developed a centralized indoor localization method using pseudo-label along with federated learning for the improved indoor localization. AdapLoc~\cite{Zhou2021AdaptiveNetworks} utilized the CNN and domain adaptation for the device-free WiFi localization in dynamic environments. In contrast, our work proposes to apply deep learning to LiDAR sensors for global localization.

\subsection{DNN-based LiDAR Odometry}
\begin{figure}
  \centering
  \includegraphics[width=\columnwidth]{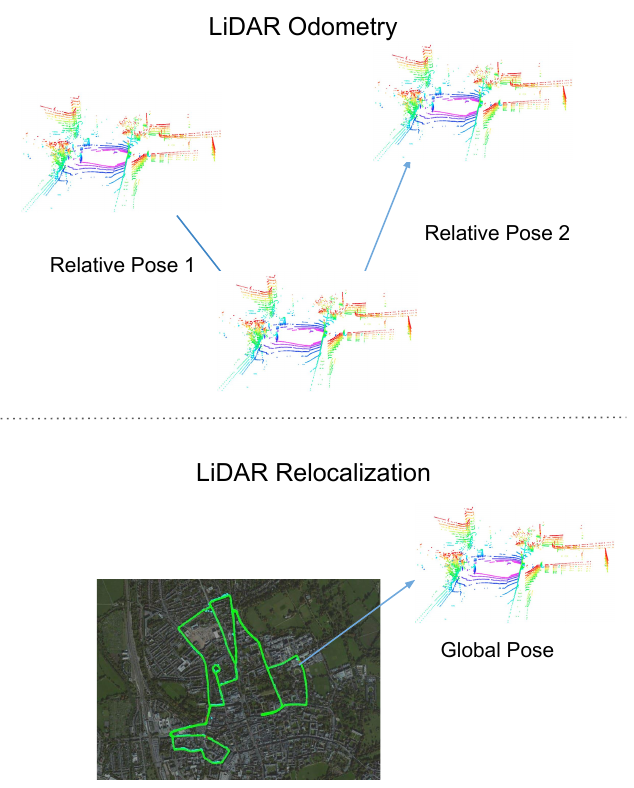}
  \caption{The difference between LiDAR odometry and LiDAR relocalization~\cite{Maddern2016,Barnes2020TheDataset}. LiDAR odometry estimates relative poses between consecutive point clouds, which produces accumulative drifts over time, while LiDAR relocalization predicts absolute poses w.r.t the world coordinate, which requires agents previously traverse scenes. These are two different tasks in localization, and this work focuses on the LiDAR relocalization.}
  \label{fig_comparisons}
\end{figure}

Recent works propose learning-based methods to estimate LiDAR Odometry, which calculates relative poses between consecutive LiDAR scans. Wang~\textit{et al.}~\cite{Wang2019DeepPCONetwork} proposed a deep parallel neural network to directly predict relative poses. Li~\textit{et al.}~\cite{Li2019DeepLocalization} developed a learning-based fusion framework with 2-D LiDAR and IMU sensors for odometry estimation. Horn~\textit{et al.}~\cite{Horn2020DeepCLR:Registration} developed a flow embedding approach to solve the fusion problem of point clouds for LiDAR Odometry. 3DFeat-Net~\cite{Yew20183DFeat-net:Registration} was developed to learn both 3-D feature detector and descriptor for point cloud matching using week supervision. Lu~\textit{et al.}~\cite{Lu2019DeepVCP:Registration} proposed a Virtual Corresponding Points method to align two point clouds accurately. Wang~\textit{et al.}~\cite{Wang2019DeepRegistration} designed novel sub-network architectures to address difficulties in the ICP method. These point cloud registration approaches can be leveraged to predict LiDAR Odometry. However, different from these methods, our work focuses on LiDAR relocalization, which estimates global poses rather than relative poses. Fig.~\ref{fig_comparisons} illustrates the difference between these two localization tasks. LiDAR odometry estimates relative poses between consecutive point clouds, while LiDAR relocalization predicts absolute poses w.r.t the world coordinate.

\subsection{Deep Learning on Point Clouds}
DNN-based feature extraction methods for point clouds \cite{Qi2017PointNet++:Space,Qi2017,Le2018,Qi2019DeepClouds} have gained significant success in recent years. VoxelNet \cite{Zhou2018VoxelNet:Detection} was developed to learn feature embeddings in voxels for object detection. PointNet++ related works \cite{Qi2017,Qi2017PointNet++:Space,Qi2019DeepClouds} have been proposed to directly process unordered point sets and learn features from points, which showed impressive performance on tasks of 3-D object detection, part segmentation, and semantic segmentation. Detailed introductions and applications of deep learning for point clouds can be found in the recent survey paper~\cite{Liu2019}.

\section{Problem Statement}
\label{sec:problem_statement}
We design a DNN-based framework for performing deep global pose regression using point cloud data from a LiDAR sensor, which is LiDAR relocalization. We predict the absolute 6-DoF poses of the mobile agent within previously visited areas. A typical use case for our method would be when a mobile agent has already visited the query places before, and then has to localize itself again when it moves across the previously-visited environment. To enable a more generic and reliable relocalization system, we only consider one LiDAR sensor input at a single timestamp rather than sequential inputs.

For each timestamp $t$,  the agent receives one point cloud frame $\mathbf{P}_{t} = \{\mathbf{x}_{i}~|~i = 1, ..., N\}$ from LiDAR, where each point $\mathbf{x}_{i}$ is a vector of describing its coordinate $(x, y, z)$. Therefore, the shape for each $\mathbf{P}_{t}$ is $(N, 3)$. The relocalization of the agent is parameterized by a 6-DoF pose $[\mathbf{t}, \mathbf{r}]^T$ with respect to the world coordinate, where $\mathbf{t} \in R^3$ is a 3-D translation vector and $\mathbf{r} \in R^4$ is a 4-D rotation vector (quaternion). To this end, deep 3-D pose regressors learn a function $\mathcal{F}$ such that $\mathcal{F}(\mathbf{P}_{t}) = (\mathbf{t}, \mathbf{r})^T$, where the function $\mathcal{F}$ is usually a neural network for DNN-based methods.

\section{Deep Point Cloud Relocalization}
\label{sec:method}
\begin{figure*}
  \centering
  \includegraphics[width=\linewidth]{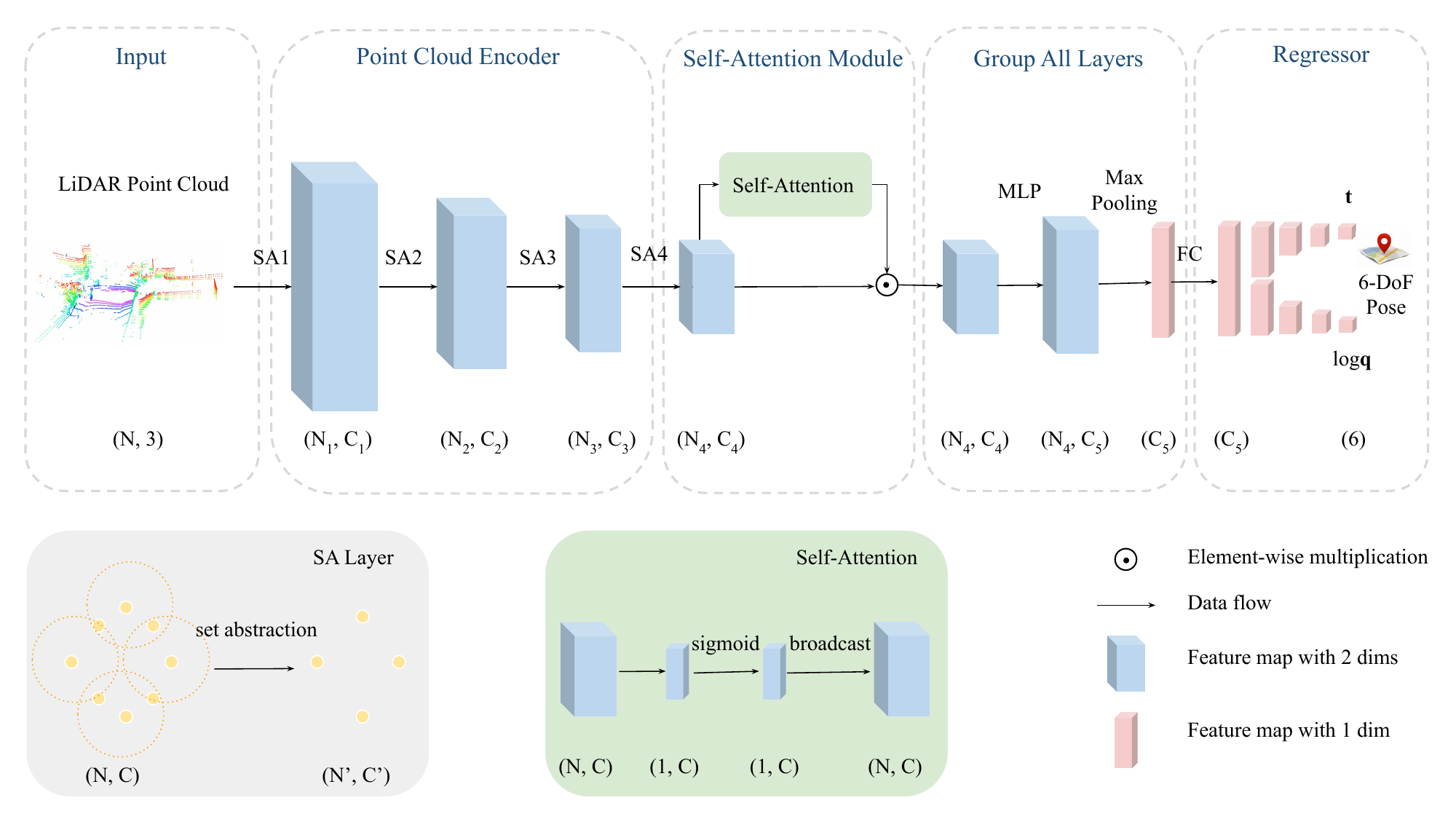}
  \caption{The architecture of PointLoc. It consists of a point cloud encoder, a self-attention module, a grouping all (GA) layers module and a pose regressor. Numbers in the parentheses represent the dimensions of feature tensors. (N, 3) represents the 3-D coordinates x, y and z of a point cloud. The encoder is composed of 4 consecutive set abstraction (SA) layers. Each SA layer shown in the diagram \cite{Liu2019FlowNet3DClouds} consists of a sampling layer, a grouping layer and a PointNet layer \cite{Qi2017}. For more details about the SA layer, please refer to PointNet++\cite{Qi2017} and FlowNet3D\cite{Liu2019FlowNet3DClouds}. The learnt point features are sent to self-attention module for eliminating the noisy features. Afterwards, these features are fed into group all (GA) layers for down-sampling to a feature vector. Finally, the pose regressor predicts the 6-DoF pose.}
  \label{PointLoc_architecture}
\end{figure*}

This section introduces our proposed PointLoc, a deep 3-D pose regressor for predicting the global pose from LiDAR sensors. The overall architecture is illustrated in Fig. \ref{PointLoc_architecture}.  
Our system consists of point cloud pre-processing, a point cloud encoder, a self-attention module, a group all layers module, and a pose regressor. The point cloud data are down-sampled to a fixed shape $(N, 3)$ as an input. The whole design is based on the PointNet-style structure, which can theoretically learn a critical subset of points for relocalization. We introduce each module individually.

\subsection{Point Cloud Pre-Processing}
The purpose of this module is to pre-process raw point clouds to fit into the neural network. Each point cloud frame of a LiDAR sensor scan contains a different number of points. However, our neural network requires the same point cloud dimensions $(N, 3)$ for its inputs. To tackle this problem, we adopt the random point cloud sampling strategy. We ensure that all the point cloud inputs have the same shape $(N, 3)$. N is set to 20,480 in this work since the average number of points in a point cloud of the Radar Robotcar Dataset\cite{Maddern2016,Barnes2020TheDataset}in our experiments is around 21,000 and we want to keep the information as much as possible.

\subsection{Point Cloud Encoder}
The goal of this module is to extract features from the point cloud. The feature representation extracted by the point cloud encoder plays a critical role in achieving accurate and reliable relocalization. Intuitively, human beings can utilize key points and features in a scene to identify where they are and conventional geometric methods are capable of performing precise localization by exploiting key points of the point cloud data.  Inspired by this, if a neural network learns a subset of key points from the original point cloud data relevant to the localization task, we can take better advantage of these key features to identify a location. 
Existing literature \cite{Qi2017,Zheng2019PointCloudMaps} has proved the critical-subset theory, i.e. for any point cloud $\mathbf{P}$, a PointNet-like structure can identify a salient point subset $\mathbf{C} \subseteq \mathbf{P}$, making it a desirable choice for our relocalization task. 

Specifically, PointNet exploits the multi-layer perceptron (MLP), feature transformation module, and max pooling layer to approximate a permutation invariant function for point cloud classification and segmentation. 
In fact, it is a universal continuous set function approximator, described as:
\begin{equation} 
\label{pointnet_eq}
	f(x_{1}, ..., x_{N}) = \phi(\textbf{MAX}\{h(x_{i})~|~x_{i} \in \mathbf{P}\})
\end{equation}
where $\phi$ and $h$ are two continuous functions (they are usually instantiated to be an MLP), and \textbf{MAX} denotes the max pooling layer \cite{Qi2017PointNet++:Space}. PointNet++ extends PointNet by recursively capturing the hierarchical features on point sets in a metric space \cite{Qi2017PointNet++:Space}. 
From the aforementioned Eq. \ref{pointnet_eq}, the result of the PointNet structure is determined by $u = \textbf{MAX}\{h(x_{i})~|~x_{i} \in \mathbf{P}\}$, and the $\textbf{MAX}$ operation takes $N$ vectors as input and outputs one vector of element-wise maximums. Thus, there exists one $x_{i} \in \mathbf{P}$ such that $u_{j} = h_{j}(x_{i})$, where $u_{j}$ is the $j^{th}$dimension of $u$, and $\mu_{j}$ is the $j^{th}$ dimension of $h(x_{i})$. These points can be aggregated into a critical subset $\mathbf{C} \subseteq \mathbf{P}$, where $\mathbf{C}$ determines $u$ and then $\phi(u)$ (more details can be found in \cite{Qi2017,Zheng2019PointCloudMaps}). Consequently, the critical-subset theory is applicable to neural networks of the structure of $\phi(\textbf{MAX}\{h(x_{i})~|~x_{i} \in \mathbf{P}\}$. The proposed PointLoc is built upon PointNet++, consisting of such a structure and thus can learn the critical subset from point clouds of LiDAR sensors in theory. 

We design our point cloud encoder based on the set abstraction (SA) layer of PointNet++ \cite{Qi2017PointNet++:Space,Qi2019DeepClouds}. The point cloud encoder is composed of 4 consecutive SA layers. Each SA layer is composed of a sampling layer, a grouping layer and a PointNet layer \cite{Qi2017}. The SA layer takes a feature matrix $\mathbf{F} \in R^{N \times C}$ as input where $N$ is the point number and $C$ is the feature dimension of each point, and outputs a feature matrix $\mathbf{F^{'}} \in R^{N^{'} \times C^{'}}$where $N^{'}$ is the sub-sampled point number and $C^{'}$ is the new feature dimension of each point (from the size ($N_{1}$, $C_{1}$) to ($N_{4}$, $C_{4}$) in Fig.~\ref{PointLoc_architecture}). We also a leverage multi-scale grouping strategy \cite{Qi2017PointNet++:Space} inside the SA layer for robust feature learning. Specifically, the layer adopts farthest point sampling to sample $N^{'}$ regions with $x_{j}$ being the region centers, and for each region with radius $r$, it extracts local features with a symmetric function as \cite{Liu2019FlowNet3DClouds}:
\begin{equation} \label{pointnet++_eq}
	\mathbf{F^{'}_{j}} = \textbf{MAX}_{\{i~|~||x_{i} - x_{j}|| \leq r\}}\{h(\mathbf{F}_{i}, x_{i} - x_{j})\}
\end{equation}
where $\mathbf{F}_{i}$ is the $i^{th}$ row of $\mathbf{F}$, $\mathbf{F}^{'}_{j}$ is the $j^{th}$ row of $\mathbf{F}^{'}$, $h: R^{C} \rightarrow R^{C^{'}}$is the MLP, and $\textbf{MAX}$ is the max pooling layer.

\subsection{Self-Attention Module}
The aim of this module is to remove outliers like moving objects from the previous extracted features for better relocalization performance. Prior works \cite{Huang2019PriorEnvironments,Wang2020AtLoc:Localization} have proved that the self-attention mechanism can improve visual sensor relocalization by removing noisy features. Therefore, we also design a neural module to automatically remove the dynamic features before regressing the final poses. Inspired by the recent works \cite{Huang2019PriorEnvironments,Wang2020AtLoc:Localization,Yang2019RobustReconstruction}, we introduce a self-attention module to learn a mask, which attempts to remove outlier features of moving objects from the original point features by conducting the element-wise dot product between the point features and the mask.

Given a set of point features $\mathbf{F} \in R^{N \times C}$ which are learned from the point cloud encoder, our attention module aims to learn a mask $\boldsymbol{M}\in R^{1 \times C}$ for the features $\mathbf{F}$. To achieve this, we use a shared MLP followed by a $sigmoid$ function to take the features $\mathbf{F}$ as input and then directly generate the mask $\boldsymbol{M}$. After that, we broadcast and mask the features $\mathbf{F}$ by $\boldsymbol{M}$, obtaining weighted features $\mathbf{\hat{F}}$ for subsequent pose regression. Specifically, since the dimensions of the point features $\mathbf{F}$ is $N \times C$ and the dimension of the learned mask $\mathbf{M}$ is $1 \times C$, we broadcast the dimensions of the mask from $1 \times C$ to $N \times C$. Afterwards, we mask the features $\mathbf{F}$ by the broadcasted mask $\mathbf{M}$ via conducting the element-wise dot product in order to remove noisy features of the point features $\mathbf{F}$. Formally, this self-attention module is defined as follows:
\begin{equation}
    \mathbf{\hat{F}} = \mathbf{F}\cdot \boldsymbol{M}
\end{equation}
where dot means element-wise dot product between $\mathbf{F}$ and $\mathbf{M}$.

\subsection{Group All Layers Module}  
 The target of this module is to aggregate features from all previous layers to generate an embedded feature vector. Specifically, shown in Fig.~\ref{PointLoc_architecture}, the input of the group all layers (GA) module is a point feature set of size $N_{4} \times C_{4}$, and then the point features are propagated to an updated point feature set of size $N_{4} \times C_{5}$ via MLP, where $C_{5}$ is larger than $C_{4}$. Next, it is down-sampled to the $C_{5}$ dimension feature vector through the max pooling layer. The embedded feature vector is then forwarded to an FC layer. After the FC layer, the $C_{5}$ dimensional feature vector is finally sent to the pose regressor for predicting the translation $\mathbf{t}$ and rotation $\mathbf{r}$ respectively. 

\subsection{Pose Regressor}  
The purpose of this module is to predict the ultimate pose. After the FC layer, the $C_{5}$ dimensional feature vector from the previous module is finally sent to the pose regressor for predicting the translation $\mathbf{t}$ and rotation $\mathbf{r}$ respectively. The pose regressor is composed of two branches of consecutive fully-connected (FC) layers. Each branch consists of 4 fully connected (FC) layers. The sizes of FC layers decrease gradually to learn features. We choose Leaky Relu as the activation function after each FC layer except for the last FC layer. The last FC layers of these two branches regress the translation and rotation separately.

\subsection{Loss Function}
Our goal is to estimate the 6-DoF pose $[\textbf{t}, \textbf{r}]^T$. Prior works \cite{Kendall2015,Kendall2017,Clark2017a, Clark2017} directly predict quaternions and use an $l_1$ or $l_2$ loss, but such a representation is over-parameterized and normalization of the output quaternion is required at the cost of worse accuracy \cite{Brahmbhatt2018}. Odometry tasks with DNNs \cite{Wang2017,Wang2019DeepPCONetwork} usually regress Euler angles, which are also not suitable here since they wrap around $2\pi$. Consequently, we employ the definition of the loss function in \cite{Brahmbhatt2018} for training our neural network, which is adapted from \cite{Kendall2017}. Given $K$ training samples $\mathcal{G} = \{\mathbf{P}_{t}~|~t = 1, ..., K\}$ and their corresponding ground-truth poses $\{[\mathbf{\hat{t}}, \mathbf{\hat{r}}]^T_{t}~|~t = 1, ..., K\}$, the parameters of the PointLoc are learned via the following loss function:
\begin{equation}
	\mathcal{L}(\mathcal{G}) =\Vert \mathbf{t}-\mathbf{\hat{t}} \Vert_{1} e^{-\beta} + \beta +
    \Vert \log \mathbf{q}-\log\mathbf{\hat{q}} \Vert_{1} e^{-\gamma} + \gamma
	\label{eq:loss}
\end{equation}
where $\beta$ and $\gamma$ are balanced factors to jointly learn translation and rotation. It is worth noting that the $\beta$ and $\gamma$ are learnable factors during training, which are initialized by $\beta^0$ and $\gamma^0$ respectively. $\mathbf{\log q}$ is the logarithmic form of a unit quaternion $\mathbf{q} = (u, \mathbf{v})$, where $u$ is a scalar and $\mathbf{v}$ is a 3-D vector. It is defined as:
\begin{equation}
    \log\mathbf{q}= \begin{cases}
            \frac{\mathbf{v}}{\Vert\mathbf{v}\Vert}\cos^{-1}u,& \text{if } \Vert\mathbf{v}\Vert \neq 0\\
            \mathbf{0},& \text{otherwise}
        \end{cases}
\end{equation}

\section{Indoor LiDAR Sensor Dataset for Relocalization}
\label{sec:vReLoc}
There is a lack of public datasets in the indoor environment with LiDAR sensors. In order to boost the research in this area, we collected a new dataset dubbed vReLoc with rich sensor modalities, e.g. vision and LiDAR sensors on a mobile robot platform. Our dataset has been released online to benefit future researchers\footnote{https://github.com/loveoxford/vReLoc}. 

The experimental robot is Turtlebot 2, mounted with a Velodyne HDL-32E LiDAR sensor and an Intel RealSense Depth Camera D435. The sensors have been carefully calibrated. The Velodyne is a lightweight pulsed laser for Detection and Ranging, which features 32 lasers across over a $40\degree$ vertical field-of-view and a $360\degree$ horizontal field-of-view. It runs at a frequency of 10Hz. Each point cloud in the dataset contains $\sim$60,000 points. The camera was employed to capture RGB images, and the size of each image is $640 \times 480 \times 3$. A Vicon Motion Tracker system is leveraged for acquiring accurate ground truth 6-DoF poses. 10 Bonita B10 cameras are used in the system, installed around the area where the dataset is collected. Each Bonita B10 has the resolution of 1 megapixel with 250 fps frame rate, and an operating range of up to 13 m. The system can track the pose of the robot at a precision of $\sim$1cm.

\begin{table}
    \footnotesize
    \centering
    \caption{\small  Dataset Descriptions on the vReLoc.}
    \vspace{.5em}
    \resizebox{\linewidth}{!}{
    \begin{tabular}{cccc}
        \toprule
        \multirow{1}{*}{Sequence}   & Scenario & Training & Test \\
        \midrule
        Seq-03       & static & \checkmark & \\
        Seq-12, Seq-15       & one-person walking & \checkmark & \\
        Seq-16       & two-persons walking & \checkmark & \\
        Seq-05, Seq-06, Seq-07 & static & & \checkmark \\
        Seq-14       & one-person walking & & \checkmark \\
        \bottomrule
    \end{tabular}
    }
    \label{tab:dataset_vReLoc}
    \vspace{-.5em}
\end{table}

The size of the Vicon room is about $4m \times 5m$. We lay out several obstacles in the scene. For the relocalization task, the scene is fixed through the whole data collection process. We utilized the Robot Operating System (ROS) for robot control and data collection. Timestamps were recorded on every frame of each sensor by the ROS, and we synchronized world timestamp across different systems from the same Network Time Protocol (NTP) server.  

A total of 18 sequences were collected of various lengths. Since the Velodyne LiDAR, RealSense camera and Vicon motion tracker system run in different frequencies, we synchronized these systems so that the image of the visual sensor and the point cloud of the LiDAR sensor in each timestamp has the same 6-DoF pose. For the static scenario, there are no moving objects in the scene. For other scenarios, there are people randomly walking in the scene. Sequences 01-10 come from the static environment, sequences 11-15 are the one-person moving scenario, and sequences 16-18 are two-persons moving scenario. In order to better represent real-world situations, in our experiments, we specially chose challenging sequences as the training dataset. 
We report our training and test sequences from the vReLoc dataset in Table~\ref{tab:dataset_vReLoc}.

\section{Experiments}
\label{sec:experiments}
In this section, we evaluate our proposed approach on the recently released outdoor Oxford Radar RobotCar \cite{Maddern2016,Barnes2020TheDataset} dataset and our proposed indoor vReLoc dataset and compare to state-of-the-art methods. 

\subsection{Implementation Details}
Adam \cite{Kingma2015} is applied to train our network with $\beta_{1} = 0.9$ and $\beta_{2} = 0.999$. We set the initial values  $\beta_{0} = 0.0$ and $\gamma_{0} = -3.0$ of the loss function following MapNet\cite{Brahmbhatt2018} and AtLoc\cite{Wang2020AtLoc:Localization}. From our experiment, if we change these values within a short range, the results remain almost the same, which is reasonable since these two parameters are learnable and they will adjust themselves to different values during training phase. The learning rate is set to 0.001, and we train 100 epochs on both datasets. For baseline image approaches, we also used data augmentation to improve the accuracy of predictions. Following the convention of existing works \cite{Kendall2015,Kendall2017,Brahmbhatt2018,Wang2020AtLoc:Localization}, we calculate the mean error for outdoor datasets and the median error for indoor datasets. 

The parameter settings of the point cloud encoder is shown in Table~\ref{tab:point_cloud_encoder}. For the parameters of point cloud encoder, especially parameters of the set abstraction layers, we employ the parameter settings from PointLoc++ and VoteNet since these settings have been demonstrated effective in point cloud-related tasks for feature extraction. In Table~\ref{tab:self_attention_module}, we present the parameter settings of the self-attention module. The parameter setting of the group all layers is shown in Table~\ref{tab:group_all_layers}. We show the parameter setting of the pose regressor in Table~\ref{tab:pose_regressor}. The neural network was trained on 2 NVIDIA TITAN V GPUs on a GPU server with 100 epochs, and it takes around 33 minutes to train the neural network for 1 epoch. The batch size is 32, and it consumes around 5.5 G of memory during training.

\begin{table}[h!]
    \footnotesize
    \centering
    \caption{\small  Parameter Setting of the Point Cloud Encoder.}
    \vspace{.5em}
    \resizebox{\linewidth}{!}{
    \begin{tabular}{ccccc}
        \toprule
        \multirow{1}{*} Layer Name  & Point Num & Radius & Sample Num & MLP \\
        \midrule
        SA1       & 2048 & 0.2 & 64 & [0, 64, 64, 128] \\
        SA2       & 1024 & 0.4 & 32 &[128, 128, 128, 256]  \\
        SA3       & 512 & 0.8 & 16 & [256, 128, 128, 256] \\
        SA4       & 256 & 1.2 & 16 & [256, 128, 128, 256] \\
        \bottomrule
    \end{tabular}
    }
    \label{tab:point_cloud_encoder}
    \vspace{-.5em}
\end{table}

\begin{table}[h!]
    \footnotesize
    \centering
    \caption{\small  Parameter Setting of the Self-Attention Module.}
    \vspace{.5em}
    \resizebox{\linewidth}{!}{
    \begin{tabular}{ccc}
        \toprule
        \multirow{1}{*} Layer Name & Point Dimension & Feature Dimension \\
        \midrule
        Self-Attention Module  & 256 & 256  \\
        \bottomrule
    \end{tabular}
    }
    \label{tab:self_attention_module}
    \vspace{-.5em}
\end{table}

\begin{table}[h!]
    \footnotesize
    \centering
    \caption{\small  Parameter Setting of the Group All Layers.}
    \vspace{.5em}
    \resizebox{\linewidth}{!}{
    \begin{tabular}{ccc}
        \toprule
        \multirow{1}{*} Layer Name (Ignore Max Pooling Layer)  & Feature Dimension\\
        \midrule
        Multi-Layer Perceptron (MLP)  & [256, 256, 512, 1024] \\
        Fully-Connected Layer (FC)  & 1024 \\
        \bottomrule
    \end{tabular}
    }
    \label{tab:group_all_layers}
    \vspace{-.5em}
\end{table}

\begin{table}[h!]
    \footnotesize
    \centering
    \caption{\small  Parameter Setting of the Pose Regressor. $\mathbf{t}$ branch and $log\mathbf{q}$ branch share the same parameter setting.}
    \vspace{.5em}
    \resizebox{\linewidth}{!}{
    \begin{tabular}{ccc}
        \toprule
        \multirow{1}{*} Layer Name & Feature Dimension  & LeakyReLu \\
        \midrule
      FC1 + LeakyReLu  & [1024, 512] & 0.2  \\
      FC2 + LeakyReLu  & [512, 128] & 0.2  \\
      FC3 + LeakyReLu  & [128, 64] & 0.2  \\
      FC4  & [64, 3] & NA  \\
        \bottomrule
    \end{tabular}
    }
    \label{tab:pose_regressor}
    \vspace{-.5em}
\end{table}

\subsection{Baselines}
To validate the performance of the proposed PointLoc, we compare it with several state-of-the-art learning-based open-source LiDAR sensor localization and visual sensor relocalization approaches. For LiDAR sensor localization approaches, we choose PointNetVLAD~\cite{Uy2018PointNetVLAD:Recognition} and Deep Closest Point (DCP)~\cite{Wang2019DeepRegistration}. PointNetVLAD is a large-scale point cloud retrieval-based approach, which can be utilized for LiDAR sensor relocalization. We create the triplet training dataset, increase the point number from 4,096 to 8,192, and set the loss margin from 0.5 to 1.0 to improve the performance, while other hyper-parameters are kept the same as the vanilla PointNetVLAD. The validation sequence FULL 5 is chosen for building up the reference database as the localization map. DCP is a DNN-based point cloud registration approach, which employs the PointNet~\cite{Qi2017} and DGCNN~\cite{Wang2018} as the embedding network. Although we deal with different tasks, we can adapt it for relocalization. Specifically, the DCP aims at the task of point cloud odometry, which demonstrates that the feature extraction design of this neural network is effective. Our task is to estimate global 6-DoF poses (relocalization) from point clouds. Therefore, we utilized the feature extraction module of DCP in our design to compare the performance. For the visual sensor relocalization baselines, we choose PoseNet17 \cite{Kendall2017} since it outperforms PoseNet and Bayesian PoseNet in previous works. AtLoc \cite{Wang2020AtLoc:Localization} is selected for comparison since it is the state-of-the-art single image-based learning approach. We also choose LSTM-Pose \cite{Walch2017} as the sequential baseline. Moreover, we also compare with MapNet \cite{Brahmbhatt2018} because it is the state-of-the-art sequential visual sensor approach. We note that sequential methods generally perform better than single image ones by utilizing time constraints. However, for the relocalization task, this past information is not always available as discussed before. We still compare with them to examine how competitive our method is. We note that we implement baseline methods and tune them for the best performance.

\begin{table}
    \footnotesize
    \centering
    \caption{\small  Dataset Descriptions on the Oxford Radar RobotCar.}
    \vspace{.5em}
    \resizebox{\linewidth}{!}{
    \begin{tabular}{cccccc}
        \toprule
        \multirow{1}{*}{Scene}   & Time &  Tag & Training & Validation & Test \\
        \midrule
        FULL1       & 2019-01-11-14-02-26 & sun & \checkmark & & \\
        FULL2       & 2019-01-14-12-05-52 & overcast & \checkmark & & \\
        FULL3       & 2019-01-14-14-48-55 & overcast & \checkmark & & \\
        FULL4       & 2019-01-18-15-20-12 & overcast & \checkmark & & \\
        FULL5       & 2019-01-15-14-24-38 & overcast & & \checkmark & \\
        FULL6       & 2019-01-10-11-46-21 & rain & & & \checkmark \\
        FULL7       & 2019-01-15-13-06-37 & overcast & & & \checkmark \\
        FULL8       & 2019-01-17-14-03-00 & sun & & & \checkmark \\
        FULL9       & 2019-01-18-14-14-42 & overcast & & & \checkmark \\
        \bottomrule
    \end{tabular}
    }
    \label{tab:dataset_radar_robotcar}
    \vspace{-.5em}
\end{table}

\subsection{Results on the Oxford Radar RobotCar}

\begin{table*}
    \footnotesize
    \centering
    \caption{\small Mean translation error (m) and rotation error (\degree) for the state-of-the-art methods on the Oxford Radar RobotCar. For PointNetVLAD and DCP, the type of input is LiDAR point clouds. The sensory data type of PoseNet17, LSTM-Pose, MapNet, and AtLoc is camera RGB image.}
    \vspace{.5em}
    \resizebox{\linewidth}{!}{
    \begin{tabular}{ccccccc|cc}
        \toprule
        \multirow{2}{*}{Scene}   & PointNetVLAD & DCP & PoseNet17 & LSTM-Pose & MapNet & AtLoc & PointLoc \\
             & \cite{Uy2018PointNetVLAD:Recognition} & \cite{Wang2019DeepRegistration} & \cite{Kendall2017} & \cite{Walch2017} & \cite{Brahmbhatt2018} & \cite{Wang2020AtLoc:Localization} & (Ours) \\
        \midrule
        FULL6      & 28.48m, 5.19\degree & 18.45m, 2.08\degree & 51.05m, 6.41\degree & 38.47m, 5.36\degree & 32.16m, 5.40\degree  & 28.57m, 7.99\degree & {\bf 13.81m, 1.53\degree} \\
        FULL7      & 17.62m, 3.95\degree & 14.84m, 2.17\degree & 80.29m, 6.51\degree & 54.59m, 4.56\degree & 47.79m, 5.44\degree  & 36.86m, 5.17\degree & {\bf 9.81m, 1.27\degree} \\
        FULL8      & 23.59m, 5.87\degree & 16.39m, 2.26\degree &111.24m, 12.78\degree &77.57m,9.74\degree &51.93m, 7.73\degree  & 58.63m, 8.53\degree & {\bf 11.51m, 1.34\degree} \\
        FULL9      & 13.71m, 2.57\degree & 13.60m, 1.86\degree &45.50m, 3.96\degree & 26.16m, 2.56\degree & 14.93m, 2.84\degree  & 10.67m, 2.88\degree  & {\bf 9.51m, 1.07\degree} \\
        \midrule
        Average    & 20.85m, 4.40\degree & 15.82m, 2.09\degree & 72.02m, 7.42\degree & 49.20m, 5.49\degree & 36.70m, 5.35\degree & 33.68m, 6.14\degree  & {\bf 11.16m, 1.30\degree}  \\
        \bottomrule
    \end{tabular}
    }
    \label{tab:result_radar_robotcar}
    \vspace{-.5em}
\end{table*}

\begin{table}
    \footnotesize
    \centering
    \caption{\small  Comparisons of the computational time and storage space of PointLoc and PointNetVLAD.}
    \vspace{.5em}
    \resizebox{\linewidth}{!}{
    \begin{tabular}{ccc}
        \toprule
        \multirow{1}{*}{}   & Computational Time &  Storage Space of the System \\
        \midrule
        PointLoc       &  0.0985 sec &  13 MB\\
        \midrule
        PointNetVLAD       &  38.0242 sec & 283 MB \\
        \bottomrule
    \end{tabular}
    }
    \label{tab:cp_pl_pv}
    \vspace{-.5em}
\end{table}

\begin{figure*}
    \centering
    \includegraphics[width=0.9\linewidth]{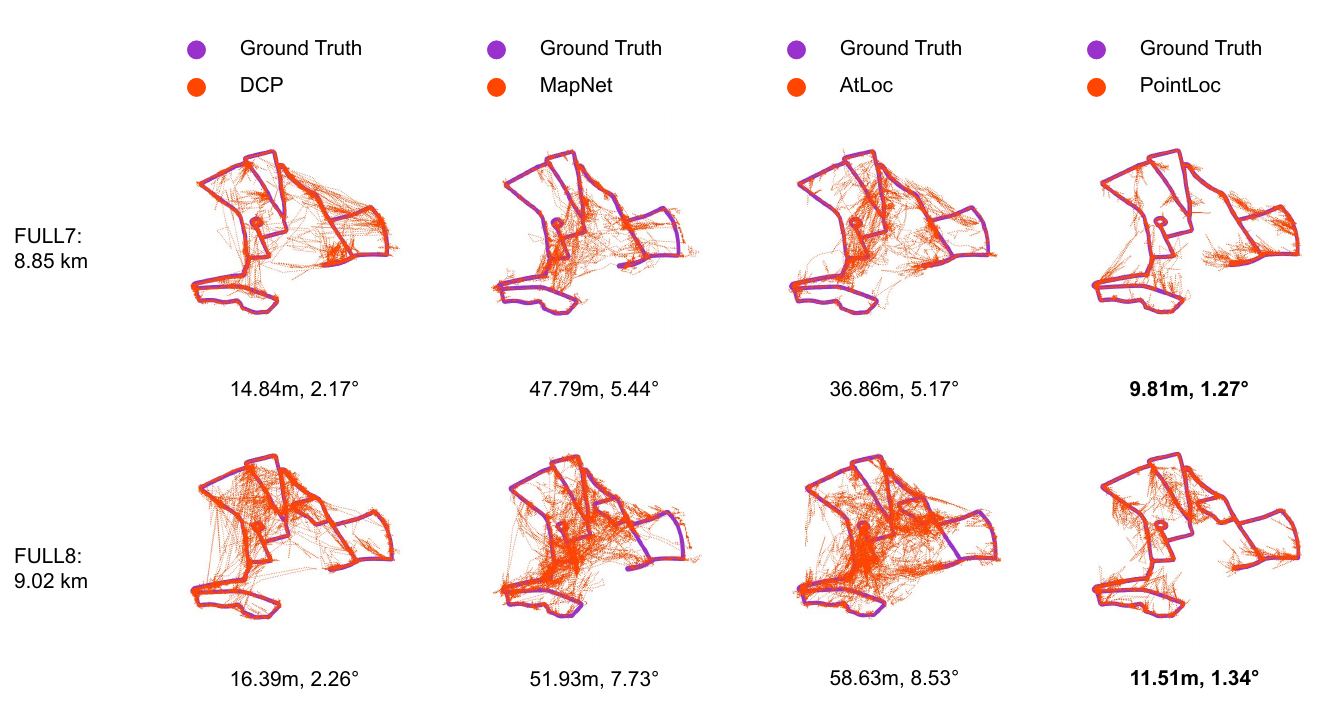}
    \caption{Trajectories of DCP, MapNet, AtLoc and the proposed PointLoc on FULL7 and FULL8 with mean translation error (m) and rotation error (\degree). The darkorchid line is the ground truth poses, and the orange-red dot line shows the estimated poses. Our PointLoc outperforms the existing LiDAR and camera relocalization approaches by a significant margin.}
    \label{fig_RobotCar}
\end{figure*}

The Oxford Radar RobotCar dataset \cite{Barnes2020TheDataset} is a radar extension to the Oxford RobotCar dataset \cite{Maddern2016}, providing data from dual Velodyne HDL-32E LiDARs and  Grasshopper2 monocular cameras. The ground truth poses are obtained by  a NovAtel SPAN-CPT ALIGN inertial and GPS navigation system (GPS/INS). 
\PAR{Dataset Description}
The data were gathered in January 2019 over thirty-two traversals of a central Oxford route, and the duration and distance of each traversal are $\sim$32mins and $\sim$9.05km respectively. The resolution of a captured RGB image is $1280 \times 960$, and each point cloud has $\sim$21,000 points. We observe that the dataset is large-scale, covers various weather conditions and has moving objects like people and cars in the scenes, all of which have significant influence on the accuracy of relocalization task, and therefore it is quite challenging. Since there is a timestamp misalignment between camera and LiDAR sensors, we synchronize timestamps with scripts and interpolate (GPS/INS) measurements to coincide with the ground truth poses. For time synchronization, the FPS (Frames per Second) of camera is 16HZ and the FPS of LiDAR is 20.02 HZ. Therefore, for every timestamp of the camera images, we collected the corresponded point cloud by searching the closest timestamp from the LiDAR point clouds. Like MapNet and AtLoc, the missing GPS/INS data is handled by interpolating values from visual odometry data which is provided by the Radar RobotCar. We report the training and test sequences we used from the Oxford Radar RobotCar in Table~\ref{tab:dataset_radar_robotcar}.

\begin{figure*}
	\centering
    \begin{subfigure}{0.225\textwidth}
        \centering
        \includegraphics[width=1.1\textwidth]{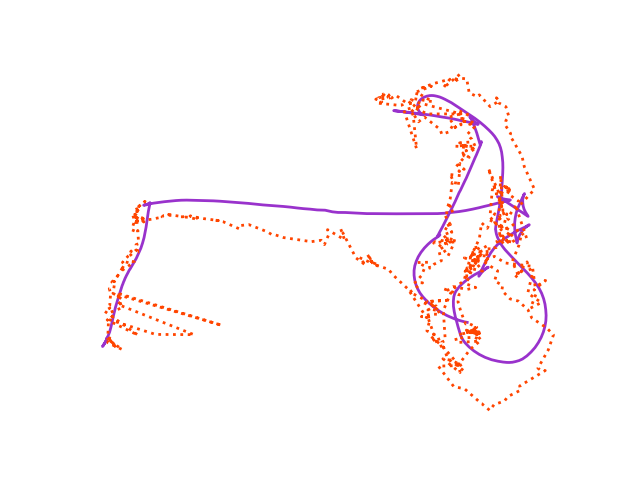}
          \caption{Seq-05 (0.12m, 3.00\degree)}
    \end{subfigure}
    \begin{subfigure}{0.225\textwidth}
        \centering
        \includegraphics[width=1.1\textwidth]{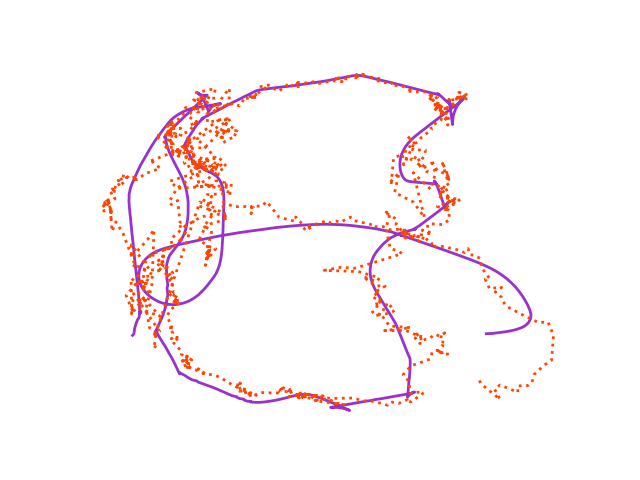}
          \caption{Seq-06 (0.10m, 2.97\degree)}
    \end{subfigure}
    \begin{subfigure}{0.225\textwidth}
        \centering
        \includegraphics[width=1.1\textwidth]{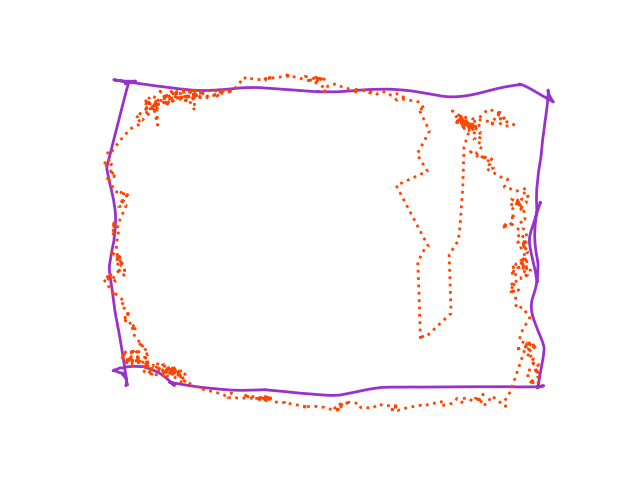}
          \caption{Seq-07 (0.13m, 3.47\degree)}
    \end{subfigure}
    \begin{subfigure}{0.225\textwidth}
        \centering
        \includegraphics[width=1.1\textwidth]{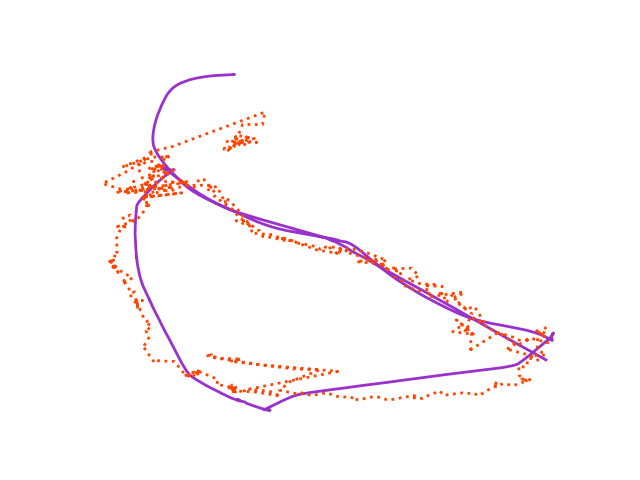}
          \caption{Seq-14 (0.11m, 2.84\degree)}
    \end{subfigure}
    \caption{Trajectories of the proposed PointLoc on the vReLoc test dataset with median translation error (m) and rotation error (\degree). The darkorchid line is the ground truth poses, and the orangered dot line is the estimated poses.}
	\label{fig_vReLoc}
\end{figure*}

\PAR{Results}
The test results of the Radar RobotCar are presented in Table~\ref{tab:result_radar_robotcar}. Following the plotting style of relocalization work \cite{Huang2019PriorEnvironments}, the trajectories of FULL7 and FULL8 of DCP, MapNet, AtLoc, and PointLoc are shown in Fig.~\ref{fig_RobotCar}. The PointLoc improves the LiDAR point cloud retrieval-based approach PointNetVLAD by 46.47\% in translation and 70.45\% in rotation, which proves the effectiveness of our proposed method. As seen from Table~\ref{tab:cp_pl_pv}, PointLoc can satisfy real-time operation and the storage space is small. The inference time is around 0.1sec, which means that by using this system, the real-time localization is
achievable at 10 Hz. This indicates that PointLoc is better than existing point cloud retrieval-based approaches for relocalization. Moreover, The PointLoc improves the DCP with PointNet by 29.46\% in translation and 37.80\% in rotation, which reveals that the proposed embedding neural network can effectively learn meaningful features for relocalization. For DCP with DGCNN, the whole neural network is difficult to train and requires large computational resources due to the large number of points and the essence of the graph architecture. Furthermore, the proposed PointLoc consistently outperforms the camera relocalization baselines by a large margin. For the best performance of deep camera relocalization, the PointLoc improves the AtLoc by 66.86\% in translation and 78.83\% in rotation. These results demonstrate that instead of utilizing RGB images as the sensory input, LiDAR point cloud can significantly improve the relocalization accuracy. Our pose predictions are even better than the sequential approaches like MapNet. In addition, the variance of learning-based camera relocalization is much larger than our approach. Therefore, the PointLoc can have stable estimation across the test dataset, which indicates that the point cloud relocalization method is more robust than the visual relocalization. 

\subsection{Results on a Real-World Indoor Robot}
We also validate our proposed PointLoc on the real-world indoor LiDAR-visual sensors dataset. Our experimental design simulates the real-world scenarios of robot movements like service robots inside a large shopping mall. The robot moved forward and backward, halting when it faces obstacles. Data was collected under three conditions: static environment, one-person walking, and two-persons walking. We named the collected dataset vReLoc since it was acquired in a Vicon room for indoor relocalization task. It includes in total 18 robot movement sequences in an indoor Vicon environment.

The median errors and trajectories of test results using our PointLoc are plotted in Fig.~\ref{fig_vReLoc}. The results demonstrated that PointLoc can be successfully applied to the real-world indoor scenarios for LiDAR sensor localization. 

\subsection{Ablation Study}
To explore the impact of different components of PointLoc, we conduct the ablation studies in Table~\ref{tab:result_ablation}. For ablation experiments, we keep all the architecture designs the same as PointLoc except that we do not contain self-attention module (w/o SA), sample 4096 points from raw point clouds (4096 Points), and utilize two fully-connected layers to predict 6-DoF poses directly (Pose Regressor). We report results on the Oxford Radar RobotCar dataset. Without self-attention module (w/o SA), the performance decreases by 7.77\% in translation and 19.75\% in rotation. This indicates that the self-attention module indeed enhances the accuracy of relocalization. Meanwhile, the PointLoc improves the same architecture with 4096 points (4096 Points) by 1.4\% in translation and 9.09\% in rotation, which demonstrates that the more sampled points can improve the performance of localization. Furthermore, the PointLoc improves the architecture with two fully-connected layers of pose regressor (Pose Regressor) by 73.83\% in translation and 84.88\% in rotation, which reveals the effectiveness of our design of Multi-Layer Perceptrons (MLPs) of two branches.

\begin{table}
    \footnotesize
    \centering
    \caption{\small Results showing the mean translation error (m) and rotation error (\degree) of ablation studies on the Oxford Radar RobotCar.}
    \vspace{.5em}
    \resizebox{\linewidth}{!}{
    \begin{tabular}{ccccc}
        \toprule
        \multirow{1}{*}{Scene}   & w/o SA & 4096 Points & Pose Regressor & PointLoc \\
        \midrule
        FULL6        & 15.24m, 1.95\degree & \textbf{13.69m}, 1.95\degree & 49.51m, 9.79\degree & \textbf{13.81m, 1.53 \degree} \\
        FULL7        & 11.32m, 1.72\degree & 10.33m, 1.40\degree & 41.53m, 8.33\degree & \textbf{9.81m, 1.27\degree} \\
        FULL8        & 13.63m, 1.80\degree & 12.35m, 1.34\degree & 42.76m, 9.26\degree & \textbf{11.51m, 1.34\degree} \\
        FULL9        & 11.03m, 1.41\degree & \textbf{8.91m}, \textbf{1.01\degree} & 36.80m, 7.01\degree & 9.51m, 1.07\degree\\
        \midrule
        Average      & 12.10m, 1.62\degree & 11.32m, 1.43\degree & 42.65m, 8.60\degree & \textbf{11.16m, 1.30\degree} \\
        \bottomrule
    \end{tabular}
    }
    \label{tab:result_ablation}
    \vspace{-.5em}
\end{table}

\section{Conclusion}
\label{sec:conclusion}
This paper presents a novel LiDAR sensor relocalization approach, PointLoc, based on deep learning. Leveraging a point-based neural network, it achieves better relocalization accuracy than previous LiDAR and visual sensor-based relocalization approaches. The approach can be applied to large-scale relocalization and robot navigation scenarios for meter-level localization requirements. It can also be leveraged in indoor environments or urban areas full of high-rise buildings as a complement when the GNSS is absent. In the future, more explorations can be done for further improving the relocalization accuracy such as eliminating the noisy point features from the point cloud or exploring intensity information for better relocalization performance.

\bibliographystyle{IEEEtran}
\bibliography{IEEEabrv,references}

\begin{IEEEbiographynophoto}{Wei Wang} is currently a PhD candidate at Department of Computer Science, University of Oxford. Before that, he obtained his Master of Science Degree from the Carnegie Mellon University and BEng Degree from the North China Electric Power University. His research interests include robotics, machine learning, and deep learning for sensory data.
\end{IEEEbiographynophoto}

\begin{IEEEbiographynophoto}{Bing Wang} is currently a PhD student at Department of Computer Science, University of Oxford. Before that, he obtained his BEng Degree at Shenzhen University, China. His research interest lies in camera localization, feature detection, description \& matching, and cross-domain representation learning.
\end{IEEEbiographynophoto}

\begin{IEEEbiographynophoto}{Peijun Zhao} is currently a PhD candidate at Department of Computer Science, University of Oxford. He has a Bachelor's degree in Computer Science and Technology from Tsinghua University. His research interests include Cyber-Physical Systems and Human-Computer Interactions.
\end{IEEEbiographynophoto}

\begin{IEEEbiographynophoto}{Changhao Chen} received his PhD degree in Department of Computer Science, University of Oxford. Before that, he obtained his MEng degree at National University of Defense Technology, China, and BEng Degree at Tongji University, China. His research interest lies in machine learning for signal processing, and intelligent sensor systems, with applications on ubiquitous localization and pedestrian navigation using mobile devices.
\end{IEEEbiographynophoto}

\begin{IEEEbiographynophoto}{Ronald Clark} is a research fellow at Imperial College London. He obtained his PhD from the University of Oxford Department of Computer Science. His work lies at the intersection of computer vision and machine learning. His research mainly focuses on allowing machines to interpret and understand the 3D world around them.
\end{IEEEbiographynophoto}

\begin{IEEEbiographynophoto}{Bo Yang} is an Assistant Professor in the Department of Computing at The Hong Kong Polytechnic University where he leads the Visual Learning and Reasoning (vLAR) Group, focusing on the fundamental research problems in machine learning, computer vision, and robotics. He completed his D.Phil/Ph.D degree (2016.10-2020.09) in the Department of Computer Science at the University of Oxford. He obtained an M.Phil degree from the University of Hong Kong and a B.Eng degree from Beijing University of Posts and Telecommunications.
\end{IEEEbiographynophoto}

\begin{IEEEbiographynophoto}{Andrew Markham} is an Associate Professor and he works on sensing systems, with applications from wildlife tracking to indoor robotics to checking that bridges are safe. He works in the cyber-physical systems group. He designs novel sensors, investigates new algorithms (increasingly deep and reinforcement learning-based), and applies these innovations to solving new problems. Previously he was an EPSRC Postdoctoral Research Fellow, working on the UnderTracker project. He obtained his Ph.D. from the University of Cape Town, South Africa, in 2008, researching the design and implementation of a wildlife tracking system using heterogeneous wireless sensor networks.
\end{IEEEbiographynophoto}

\begin{IEEEbiographynophoto}{Niki Trigoni} is a Professor at the Oxford University Department of Computer Science and a fellow of Kellogg College. She obtained her DPhil at the University of Cambridge (2001), became a postdoctoral researcher at Cornell University (2002-2004), and a Lecturer at Birkbeck College (2004-2007). At Oxford, she is currently Director of the EPSRC Centre for Doctoral Training on Autonomous Intelligent Machines and Systems, a program that combines machine learning, robotics, sensor systems and verification/control. She also leads the Cyber-Physical Systems Group, which is focusing on intelligent and autonomous sensor systems with applications in positioning, healthcare, environmental monitoring and smart cities.
\end{IEEEbiographynophoto}

\end{document}